\pdfoutput=1
\documentclass[journal]{IEEEtran}
\usepackage{cite}

\ifCLASSINFOpdf
\else
\fi

\usepackage{amsfonts}
\usepackage{multirow}
\usepackage{pgfplots}
\usepackage{mathtools}
\usepackage{threeparttable}

\usepackage{enumerate}
\usepackage{color,xcolor,colortbl}
\usepackage[switch]{lineno}
\usepackage{times}
\usepackage{epsfig}
\usepackage{graphicx}
\usepackage{amsmath}
\usepackage{amssymb}
\usepackage{multirow}
\usepackage{pifont}

\usepackage{algorithm}  
\usepackage{algpseudocode}

\begin{document}

%
\title{Fully and Weakly Supervised Referring Expression Segmentation with End-to-End Learning}
%
%
%

\author{Hui~Li,
     Mingjie~Sun,   Jimin~Xiao,
    Eng~Gee~Lim,
    and~Yao~Zhao,

}

\maketitle

\begin{abstract}

Referring Expression Segmentation (RES), which is aimed at localizing and segmenting the target according to the given language expression, has drawn increasing attention. Existing methods jointly consider the localization and segmentation steps, which rely on the fused visual and linguistic features for both steps. We argue that the conflict between the purpose of identifying an object and generating a mask limits the RES performance. To solve this problem, we propose a parallel position-kernel-segmentation pipeline to better isolate and then interact the localization and segmentation steps. In our pipeline, linguistic information will not directly contaminate the visual feature for segmentation. Specifically, the localization step localizes the target object in the image based on the referring expression, and then the visual kernel obtained from the localization step guides the segmentation step. This pipeline also enables us to train RES in a weakly-supervised way, where the pixel-level segmentation labels are replaced by click annotations on center and corner points. The position head is fully-supervised and trained with the click annotations as supervision, and the segmentation head is trained with weakly-supervised segmentation losses. To validate our framework
on a weakly-supervised setting, we annotated three RES benchmark datasets (RefCOCO, RefCOCO+ and RefCOCOg) with click annotations.  
Our method is simple but surprisingly effective, outperforming all previous state-of-the-art RES methods on fully- and weakly-supervised settings by a large margin. The benchmark code and datasets will be released.
\end{abstract}

\begin{IEEEkeywords}
Referring Expression Segmentation, Weakly-Supervised, End-to-End, Position-Kernel-Segmentation.
\end{IEEEkeywords}

%
\IEEEpeerreviewmaketitle

%
%
%
%
\section{Introduction}
\label{sec:intro}

\begin{figure}
\begin{center}
\includegraphics[width=0.9\linewidth]{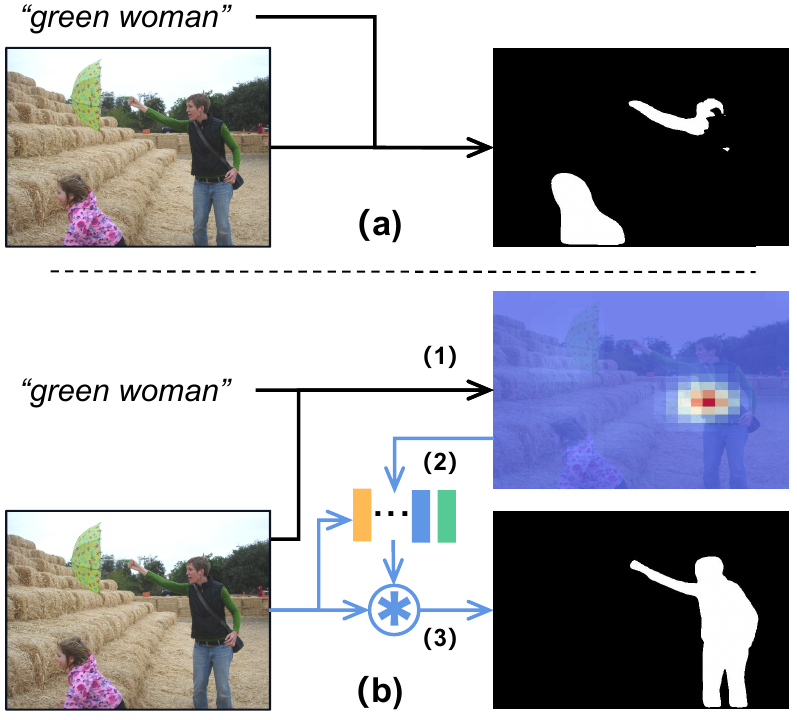}
\end{center}
   \caption{Comparison of our method with the vanilla method for referring expression segmentation, using the RefCOCO dataset as an example. (a) \textbf{The vanilla method} predicts the mask of the target object via the fused feature of the vision and language. (b) \textbf{Our method} predicts the mask via a position-kernel-segmentation process, where linguistic
information will not directly contaminate the visual feature for segmentation. Note that the blue lines are the main difference from previous works.}
\label{fig:intro}
\end{figure}

\IEEEPARstart{R}{eferring} Expression Segmentation (RES)~\cite{liu2017recurrent,kamath2021mdetr,yang2021bottom,vision-language-transformer, feng2021encoder,jing2021locate} is aimed at generating the segmentation mask on the most relevant object referring to the given language expression. RES has wide applications in human-robot interactions, such as robotic navigation with language instructions~\cite{anderson2018vision,ning2020polar,9265290}. However,
RES is still a challenging task due to the complex interactions between the two modalities.

The RES segmentation quality is affected by both target object localization and segmentation~\cite{jing2021locate}. Target object localization is aimed at identifying the preliminary target location by jointly analysing language expression and visual content. Target object segmentation compares the localized target region with surrounding regions to draw the object contour as the segmentation mask. 

Some early RES works~\cite{kamath2021mdetr,vision-language-transformer,yang2021bottom} do not separate the localization and segmentation modules. The segmentation masks are directly generated without intermediate localization results.
Recently, end-to-end pipelines, which locate the target object and then segment it, have been introduced~\cite{jing2021locate,LI202299}. These pipelines decouple the referring image segmentation task into two sequential tasks, and explicitly locate the referred object guided by language expression. The two-step structure uses localization as relevance filtering, leading to more accurate segmentation masks \cite{LI202299}.
Other works use semantic segmentation tools to improve RES quality. For instance, object boundaries are utilized to strengthen visual features~\cite{feng2021encoder}, or Conditional Random Field (CRF) is used for postprocessing~\cite{liu2017recurrent,li2018referring,ye2019cross,huang2020referring,hu2020bi}. 

However, the difference between localization and segmentation has not been thoroughly considered in previous methods. Specifically, for localization, its primary purpose is to distinguish the target object from other objects, where the language expression is essential for excluding nontarget objects. As shown in Fig.\ref{fig:intro}(a), for the query ``green woman'', the target in the image is the lady in green clothes but blue trousers. To predict the preliminary target location, ``green'' contributes to distinguishing the target person from other persons in different clothes. However, the incomplete information in language expressions may distract the segmentation process. The mismatch between the ``green'' clothes and the target's blue trousers may confuse the segmentation module with the trousers part unmasked.


To solve this issue, we design an end-to-end parallel position-kernel-segmentation pipeline, where the position prediction can be isolated from the segmentation module, as shown in Fig.\ref{fig:intro}(b). Specifically, the position head (1) finds the target object's preliminary region according to the given image and query sentence. The kernel head (2) selects the kernels corresponding to the visual feature points within the target region, predicted from the position module. The segmentation head (3) generates the final fine-grained mask, where the selected kernels are used to conduct the convolutional operation. For example in Fig.\ref{fig:intro}, our position head predicts the center position of the ``green woman'' via the probability map of the entire image. Then, our kernel head selects the most relevant regions. Ultimately, the segmentation head predicts the final mask of that woman (dismiss ``left girl'') based on the selected regions. Due to the independent structure, the localization branch is isolated from the segmentation branch. The segmentation branch uses the visual features and kernels as input, solely focusing on generating the segmentation mask. 

Inspired by weakly supervised semantic segmentation~\cite{lin2016scribblesup,zhang2020reliability,zhang2021affinity} and weakly supervised salient detection~\cite{tian2020weakly,yu2021structure}, this paper also presents the first attempt on Weakly supervised Referring Expression Segmentation (WRES). We provide annotations on three main RES datasets (RefCOCO, RefCOCOg and RefCOCO+), where pixel-level segmentation labels are replaced by click annotations on the target object center and corner points. Compared with REG, WRES dramatically reduces the annotation cost of pixel-level labeling.

Specifically, our parallel position-kernel-segmentation pipeline fits the WRES task well, which could be treated as a benchmark method. (1) The position head is fully-supervised and trained with the click annotations as supervision. Such a design is not applicable to other RES methods, as they do not rely on object localization information as supervision. (2) The segmentation head is trained with partial cross-entropy loss and augmented with CRF loss, as full pixel-level segmentation labels are not available.


In summary, the contributions of this paper are listed as follows: 

\begin{itemize}

\item  We propose a position-kernel-segmentation framework for RES, where localization and segmentation are separated. In the localization branch, the linguistic feature is fused into the visual feature for better localization. The segmentation branch uses the visual feature and kernels as input, solely focusing on generating better segmentation masks. 

\item This position-kernel-segmentation framework is naturally suited for weakly-supervised training, with object center and corner clicks as labels.
To validate our framework on the weakly-supervised setting, we annotated three benchmark datasets for the weakly-supervised RES using click annotations, which will be released to the public. 
 
\item For the fully-supervised setting, we achieve state-of-the-art (SOTA) performances on three primary datasets on the RES task, including RefCOCO~\cite{yu2016modeling}, RefCOCO+~\cite{yu2016modeling} and RefCOCOg~\cite{mao2016generation,nagaraja2016modeling}, where the average mIoU scores (val/testA/testB) are significantly improved from the previous SOTA method by 3.07\%, 4.66\% and 2.39\%, respectively. For the weakly-supervised setting, our method also obtains satisfactory performances, with average mIoU scores of 49.27\%, 37.79\% and 40.98\%, respectively. 

\end{itemize}

\begin{figure*}
\begin{center}
\includegraphics[width=\linewidth]{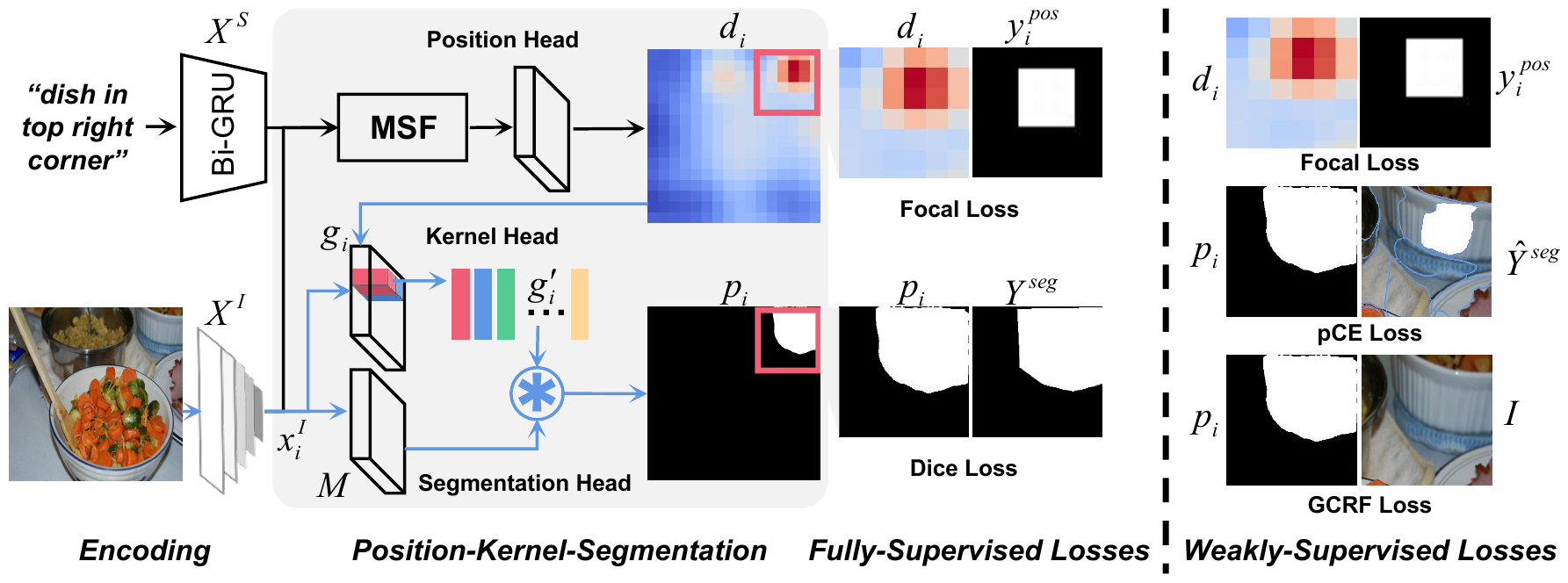}
\end{center}
   \caption{Overview of the proposed position-kernel-segmentation RES pipeline. The main parts of the proposed method consist of three heads: (a) \textbf{The position head} predicts a heatmap to localize the object described by the referring expression via a Multi-Step Fusion (MSF) process. (b) \textbf{The kernel head} selects the kernel vectors from the image feature map using the position information. (c) \textbf{The segmentation head} predicts the final segmentation mask of the target object. The MSF module fuses the linguistic and visual features. Best viewed in color.}
\label{fig:method}
\end{figure*}

\section{Related Work}

\subsection{Vision and Language}
Multimedia systems have drawn increasing attention to big data artificial intelligence. Yang \textit{et al. }~\cite{yang2021multiple} first introduced multiple knowledge representation (MKR), where the knowledge representation learns from different abstraction levels, different sources, and different perspectives for artificial intelligence.

The vision-language task has a variety of real-world applications, including Visual Question Answering (VQA)~\cite{antol2015vqa,Jiang_2020_CVPR,le2020hierarchical,bansal2020visual,Chen_2020_CVPR,kim2020modality,wang2020general}, Image-Text Matching (ITM)~\cite{karpathy2015deep,wang2019camp,chen2020imram, Su2020VL-BERT,lu2019vilbert,tan2019lxmert,8540429,9406055,9222079}, Referring Expression Grounding (REG)~\cite{rohrbach2016grounding, yu2018mattnet, yang2019fast,Liu_2019_CVPR, sun2021discriminative, kamath2021mdetr}, Referring Video Object Segmentation (RVOS)~\cite{khoreva2018video,seo2020urvos}, \emph{etc}. An important line of research in existing vision-language works is to align the image visual feature with the text linguistic feature~\cite{Wang2020CVSE,kipf2016semi,chen2020imram,9349150}. For the image-text matching task, Wang \textit{et al.} \cite{Wang2020CVSE} proposed consensus-aware embedding to enhance visual and language features, using the statistical correlations of words via a GCN~\cite{kipf2016semi}. The fused representation takes into account the cross-modality interactions via different fusion methods. Chen \textit{et al.} \cite{chen2020imram} propose an iterative matching strategy with recurrent attention memory to associate semantic concepts between visual features and linguistic features from low-level objects to higher-level relationships. For the action recognition task, Zhu \textit{et al.} \cite{9349150} proposed cross-layer attention, which learns the importance weight of different feature layers.

As an important task for human-robot interactions, REG is aimed at localizing a target object in an image described by a referring expression. Yu \textit{et al. }~\cite{yu2018mattnet} view the proposal-based method as a region-retrieval problem with proposals of all candidate objects provided in advance. Yang \textit{et al. }~\cite{yang2019fast} combine the linguistic feature of the query sentence and the visual feature from the image to generate a multimodality feature map. Then, the binary classification scores (foreground or background) for all points and their corresponding bounding box coordinates are predicted. 

\subsection{Referring Expression Segmentation}

RES is aimed at segmenting the objects based on the expression instead of at generating their bounding boxes, which request fine-grained masks with more details, making it more challenging.

Early RES works focused on directly improving the segmentation quality, and never considered the separation between the localization modules and the segmentation modules. The transformer-based architecture is adopted in \cite{kamath2021mdetr} to better fuse the vision and language features. Multiple sets of textual attention maps are proposed in~\cite{vision-language-transformer}, and in each attention map, an attention weight is predicted for every single word in the query sentence. Thus, different attention maps can represent diversified textual comprehensions from different aspects to better understand the query sentence. Different from the aforementioned works that combine localization and segmentation modules, Jing \textit{et al.}~\cite{jing2021locate} first proposed an end-to-end locate-then-segment pipeline, where the localization result is concatenated with the fused visual-linguistic feature for segmentation prediction. CMS-Net \cite{LI202299} proposes an end-to-end cross-modality synergy network, which uses the language as guidance to get the language-aware visual feature, and the language-aware visual feature is fused with the original language feature for final segmentation predictions.

Other works use semantic segmentation tools to improve RES quality. Feng \textit{et al.}~\cite{feng2021encoder} use the boundary information as the additional supervision in the training process, which strengthens the visual features to generate smoother masks. Hu \textit{et al.}~\cite{hu2020bi} first introduced the widely-employed segmentation tool, CRF, to refine the output semantic segmentation map, which is also adopted by later works~\cite{liu2017recurrent,li2018referring,ye2019cross,huang2020referring,hu2020bi}. 

Unlike existing methods, this is the first work to design an end-to-end position-kernel-segmentation pipeline that separates the localization and segmentation modules, where linguistic information will not directly contaminate the visual feature for segmentation. Different from previous methods, our position head influences the segmentation head via the kernel head, where the noise of the attention does not directly influence the segmentation process. However, the previous method does not isolate the segmentation branch,  thus the unstable attention in localization affects the mask generation process.

\subsection{Semantic Segmentation}

Supervised semantic segmentation takes pixel-level labels as supervision to train a model to predict the precise object masks. 
The recent deep-learning-based methods use anchor-based (Mask R-CNN~\cite{he2017mask}), kernel-based (SOLOv2~\cite{wang2020solov2}), and transformer-based (Swin Transformer~\cite{liu2021swin}) solutions for precise mask generation. However, supervised semantic segmentation has one major limitation: pixel-level annotation is time-consuming and costly. 

To reduce the human annotation cost, many works investigate weakly-supervised semantic segmentation including image-level~\cite{wei2017object}, scribble-level~\cite{zhang2020reliability}, and point-level~\cite{zhang2021affinity} annotations. For image-level labels, the basic idea is to use the classification information to generate Class Activation Map (CAM), where the activated areas can be treated as the seeds for pseudo labels. For scribble- and point-based methods with partial pixel-level annotations, ScribbleSup~\cite{lin2016scribblesup} uses the superpixel method of Simple Linear Iterative Clustering (SLIC)~\cite{7885581} to expand the original labels. Additionally, other studies focus on reducing the noise of the expanded pseudo labels. For instance, Tang \textit{et al. }\cite{tang2018normalized} use the normalized cut loss to cut noisy labels lower than the trust threshold. 

Weakly-supervised RES is more challenging than weakly-supervised semantic segmentation, since weakly-supervised semantic segmentation has classification information to generate a CAM as pseudo labels, which is not available in weakly-supervised RES. In our setting, the only labels for weakly-supervised RES are the object center and corner points.


\section{Methodology}
As shown in~Fig.\ref{fig:method}, our framework mainly consists of 5 components: the feature encoding module, position head, kernel head, segmentation head and training losses. The position head, kernel head and segmentation head follow a position-kernel-segmentation workflow to generate the segmentation mask of the target object.

\begin{itemize}
\item  \emph{The feature encoding module} encodes input images and referring sentences into high dimensional features (Sect.\ref{sec:init}). All levels of the hierarchical visual features will be applied in the following steps.

\item \emph{The position head} fuses linguistic and visual features via our designed MSF mechanism; then, the position heatmap is predicted, describing the target object's spatial position in the image (Sect.\ref{sec:position}). 

\item \emph{The kernel head} takes the visual features and position heatmaps to select the kernel features, where the ultimate selections correspond to the high probability regions from the position heatmaps (Sect.\ref{sec:kernel}). 

\item \emph{The segmentation head} uses the visual features and selected kernels to predict the final segmentation mask of the target object (Sect.\ref{sec:seg}).

\item  \emph{The fully-supervised setting} uses two parallel losses for target object localization and segmentation (Sect.\ref{sec:loss}); the \emph{weakly-supervised setting} uses three losses, with an addition loss considering color consistency (Sect.\ref{sec:loss_w}).


\end{itemize} 

\subsection{Feature Encoding Module}
\label{sec:init}
For an input image $I\in\mathbb{R}^{H\times W\times 3}$ with height $H$ and width $W$, the initial visual pyramid features are defined as $X^I$=$\{x^I_i\}_{i=1}^{L}$ with $L$ levels, obtained from the pretrained feature extractor Feature Pyramid Network (FPN)~\cite{lin2017feature}, where $x^I_i$ corresponds to the $i$-th level of the pyramid features.


For a sentence with $N$ words $S$=$\{u_t\}_{t=1}^{N}$, its linguistic feature is denoted as $X^S$=$\{x^S_t\}_{t=1}^{N}$, where each word's feature $x^S_t$ at position $t$ is generated by the linguistic feature extractor Glove~\cite{pennington2014glove}, followed by the sequential bidirectional feature extractor Bi-GRU~\cite{cho2014properties}.

\subsection{Position Head}
\label{sec:position}

Our position head predicts heatmaps to localize the target object described by the referring expression via a MSF process. Using the visual feature $X^I$ and linguistic feature $X^S$ from Sect.\ref{sec:init} as inputs, the position head predicts the heatmaps $D$=$\{d_i\}_{i=1}^{L}$, which have the same levels $L$ as $X^I$. $d_i$ is the $i$-th position heatmap corresponding to feature $x^I_i$. Each pixel value in $d_i\in\mathbb{R}^{h_i\times w_i}$ represents the probability of whether this pixel belongs to the target object. 

\textbf{Multi-Step Fusion:}
\label{sec:MS}
The visual features $X^I$ are fused with the linguistic feature $X^S$ to obtain the position heatmaps $D$. Specifically, $d_i$ is obtained after fusing $x^I_i$ with $X^S$. 

A MSF process is conducted to fuse the visual and language features, following one fully convolutional layer to calculate the probability values in the position heatmap.

The MSF takes $J$ steps. In each step, the linguistic feature $X^S$ is blended into feature $F_i^j$ (where $F_i^0=x^I_i$) to obtain the next step fused feature $F_i^{j+1}$:
\begin{equation}
\label{eq:MS1}
F_i^{j+1} = \mathit{Fusion}(X^S,F_i^j), 
0 \leq j \leq J-1 ,
\end{equation}
where $F_i^j\in\mathbb{R}^{h_i\times w_i\times c_i^j}$ is the fused feature on the $i$-{th} level and at the $j$-th step, and $J$ is the total step number.

\begin{figure}
\begin{center}
\includegraphics[width=1\linewidth]{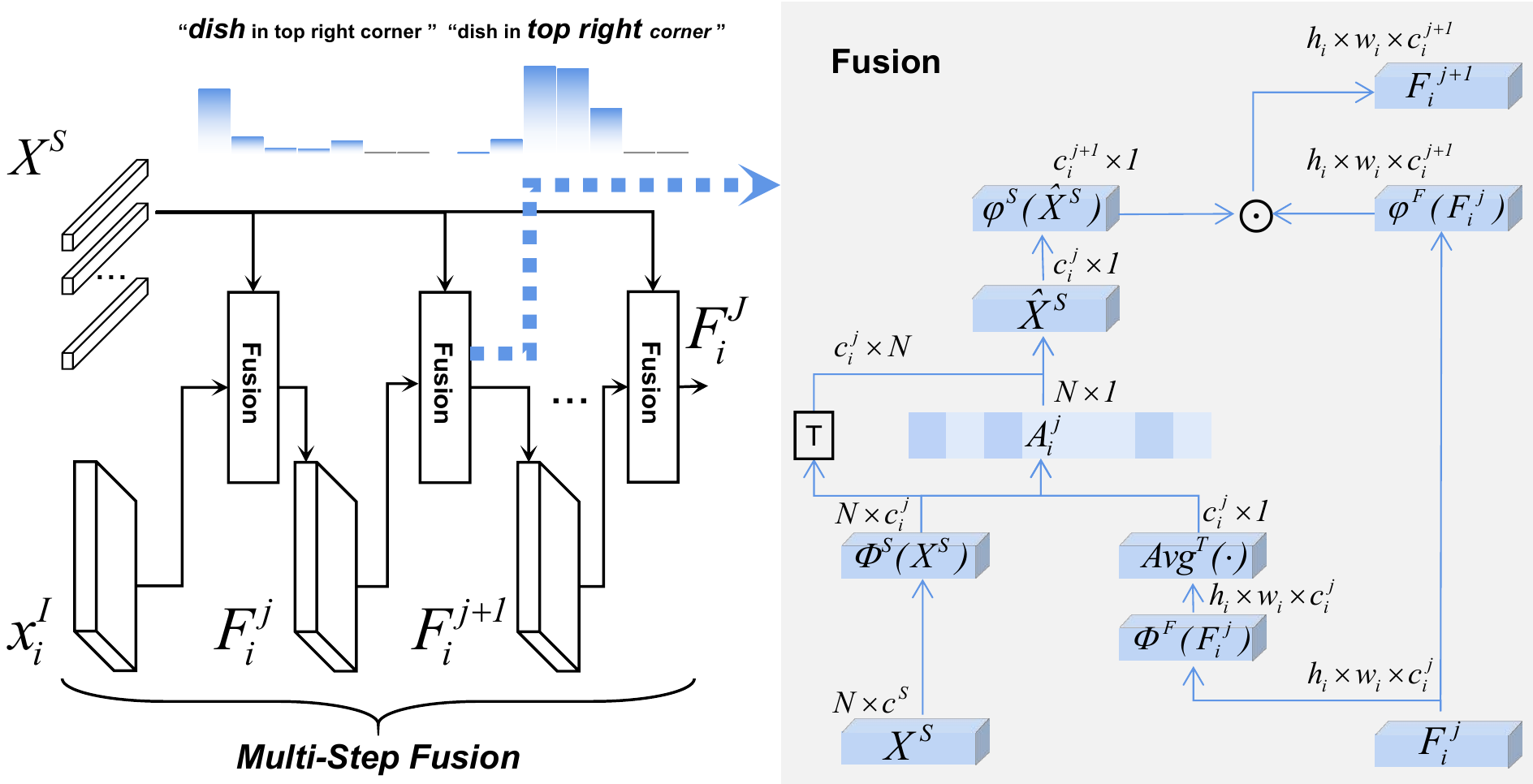}
\end{center}
   \caption{The MSF (left) and the details of fusion with attention on level $i$ and step $j$ (right). The fusion process generates the next step fused feature $F_i^{j+1}$ provided with $X^S$ and $F_i^{j}$. The feature size of each element is marked above each feature. $\odot$ represents the element-wise multiplication on the channel dimension, where $\varphi^S(\hat{X}^S)$ is broadcast to the same size of $\varphi^F(F_i^j)$. $T$ is the transpose operation.}
\label{fig:att}
\end{figure}

\begin{figure}
\centering
\includegraphics[width=0.95\linewidth]{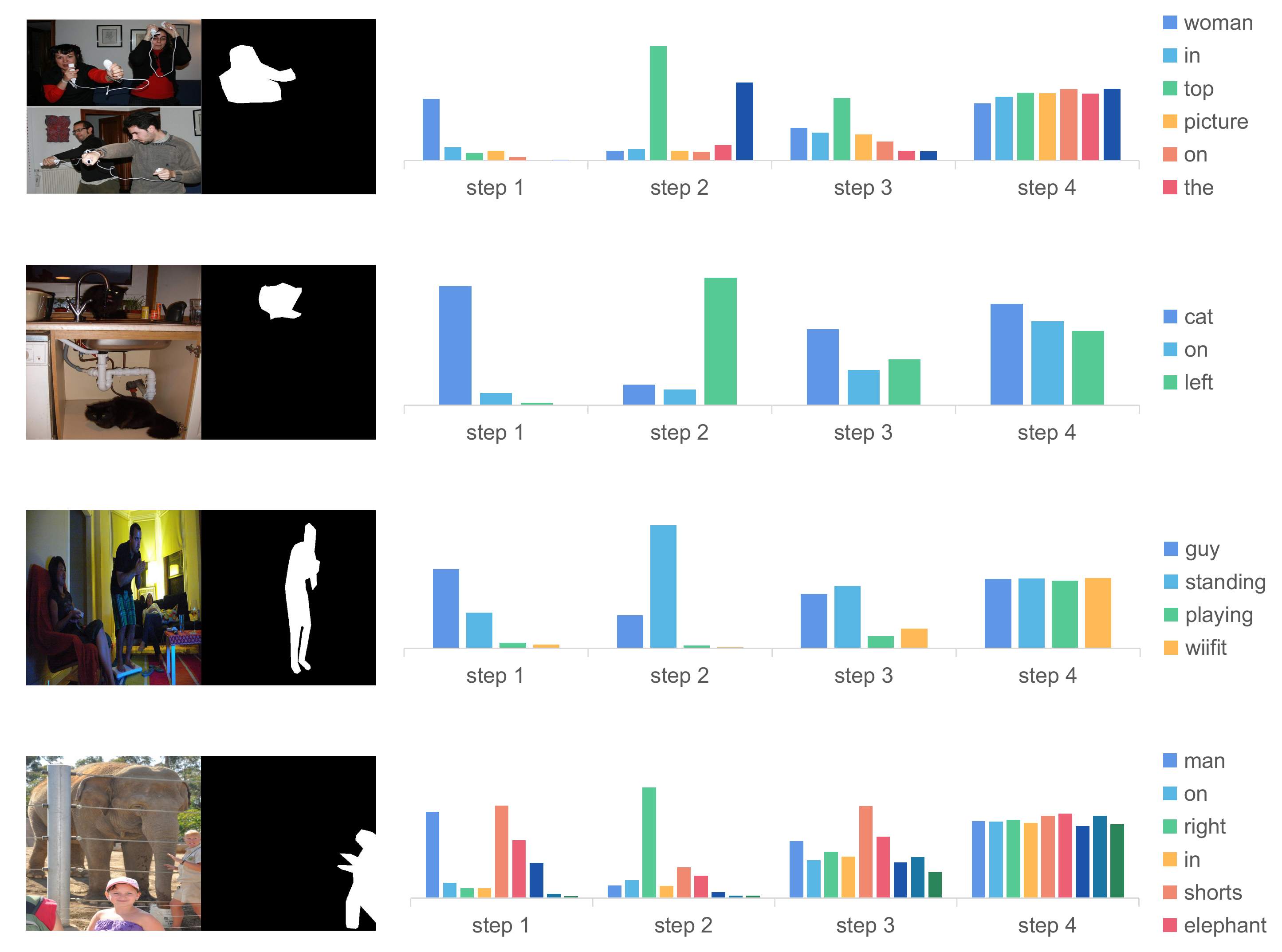}
   \caption{Examples of the different attention values that are allocated to different phrases in different fusion steps.}
\label{fig:att_word}
\end{figure}

In different fusion steps, different attention values are allocated to different phrases to focus on different phrases. For instance, ``dish'' takes the highest attention value in the first fusion step while the second step finds ``top'' and ``right'' to be the most relevant, as shown in Fig.\ref{fig:att} (left). More examples are provided be found in Fig.\ref{fig:att_word}.

Using the final fusion feature $F_i^J$ as input, one Fully Convolutional Layer (FCL) is utilized to predict the final position heatmap $d_i$:
\begin{equation}
\label{eq:MS2}
d_i = \mathit{FCL}(F_i^J).
\end{equation}

\textbf{Fusion with Attention:} 
This subsection describes the implementation of~Eq.(\ref{eq:MS1}) using the attention mechanism.
Specifically, the detailed process, which consists of an attention operation between the language feature and the visual feature, is illustrated in~Fig.\ref{fig:att} (right).

The language attention calculates the importance of each word in the sentences based on the image visual feature. The language attention matrix for the $i$-th level and $j$-th step, \emph{i.e.,} $A_i^j \in\mathbb{R}^{N \times 1}$, is calculated by:
\begin{equation}
\label{eq:att1}
 A_i^j = \phi^S(X^S) (\mathit{Avg}(\phi^F(F_i^j)))^T,
\end{equation}
where $\phi^S(\cdot)$ and $\phi^F(\cdot)$ are convolutional operations to unify the linguistic feature and fused features into the same channel size. Their final sizes are $N\times c_i^j$ and $h_i\times w_i\times c_i^j$. To represent the entire image, we use $\mathit{Avg}(\phi^F(F_i^j)) \in\mathbb{R}^{1 \times c_i^j} $, which is the average feature of all feature pixels in the $h_i\times w_i$ map.  
$(\cdot)^T$ is the transpose operation. Then, matrix multiplication is performed to calculate the final attention values $A_i^j$.

With the language attention matrix $A_i^j$ in Eq.(\ref{eq:att1}), we obtain the weighted average linguistic feature $\hat{X}^S$=$(\phi^S(X^{S}))^T Sigmoid(A_i^j)$, where the sigmoid function is used to scale the attention values to range $(0,1)$.


The fused feature $F_i^{j+1}$ is calculated as:
\begin{equation}
\label{eq:att2}
\begin{aligned}
 F_i^{j+1} = \varphi^S(\hat{X}^S) \odot \varphi^F(F_i^j),
\end{aligned}
\end{equation}
where $\varphi^S(\cdot)$ and $\varphi^F(\cdot)$ are convolutional operations to convert the weighted linguistic feature and fused feature into the next step's channel size by one convolutional layer. The final fused feature is calculated by element-wise multiplication. Specifically, $\varphi^S(\hat{X}^S)$ is broadcast to the same size of $\varphi^F(F_i^j)$. Note that Sect.\ref{sec:position} is the only module where the language information is implemented.


\subsection{Kernel Head}
\label{sec:kernel}

The kernel head uses the visual features $X^I$ and the position heatmaps $D$ to select the kernels $G'$, which are used in the segmentation head to produce the target object mask. 


Specifically, the first step in our kernel head is to generate the kernel map pools $G$=$\{g_i\}_{i=1}^{L}$, using the image features $X^I$. $g_i$ corresponds to the $i$-th feature $x^I_i$, which is calculated as
\begin{equation}
\label{eq:G}
g_i = \mathit{CoordConv}(x^I_i).
\end{equation}
$\mathit{CoordConv}(\cdot)$~\cite{liu2018intriguing} represents a set of convolutional operations on the coordinate-enhanced visual feature. Each $c_{i}$-channel feature point in $x^I_i$ (resolution is $h_i\times w_i$) is concatenated with its two-dimensional coordinates, where the channel number becomes $c_{i}+2$.

The final $g_i\in\mathbb{R}^{h_i\times w_i\times c'}$ shares the same height $h_i$ and width $w_i$ with the position heatmap $d_i$. Based on the confidence values in $d_i$, $k_i$ kernel features $g'_i\in\mathbb{R}^{k_i\times c'}$ are selected from $g_i$ for level $i$:
\begin{equation}
\label{eq:filter}
g'_i = \big\{g_i[m,n,:]\big|d_i[m,n] > \lambda_{f} \big\}_{m=1,n=1}^{h_i, w_i},
\end{equation}
where $[m,n]$ is the coordinate within $h_i\times w_i$. $\lambda_{f}$ is the threshold, where the probability value $d_i[m,n]$ is higher than $\lambda_{f}$ for the selected kernel position $[m,n]$. The element number of the selected kernel set $G'$=$\{g'_i\}_{i=1}^{L}$ is $\sum_{i=1}^{L} k_i$. Note that, during the training process, the kernels $g'_i$ are selected by the ground-truth label from the position head. The details of generating the ground-truth label for the position head are introduced later in Sect.\ref{sec:loss}.

\subsection{Segmentation Head}
\label{sec:seg}

The segmentation head uses the visual features $X^I$ and the selected kernels $G'$ to predict the final segmentation mask $P$ of the target object.

First, a segmentation decoder~\cite{kirillov2019panoptic} is employed to generate the mask feature map $M$ by taking the $L-1$ levels of the visual feature pyramid $X^I$ as input:
\begin{equation}
\centering
\label{eq:mask_f}
M = \mathit{Decoder}(\{x^I_i\}^{L}_{i=2}), 
\end{equation}
where $M\in\mathbb{R}^{\frac{H}{4}\times \frac{W}{4}\times c'}$ has a quarter size of the original input image $I$ and the same channel number $c'$ as $g'_i$ in Eq.(\ref{eq:filter}). 

Second, the final predicted mask $P$=$\{p_i\}_{i=1}^{L}$ (each $p_i\in\mathbb{R}^{k_i\times \frac{H}{4}\times \frac{W}{4}}$) is generated using the convolution operation between the mask feature $M$ and the kernel $g'_i \in G'$,
\begin{equation}
\centering
\label{eq:mask}
p_i = \mathit{Sigmoid}(\mathit{conv}(M, g'_i)), 
\end{equation}
where the sigmoid operation is used to scale the predicted values to range $(0,1)$. 

\begin{figure}
\begin{center}
\includegraphics[width=\linewidth]{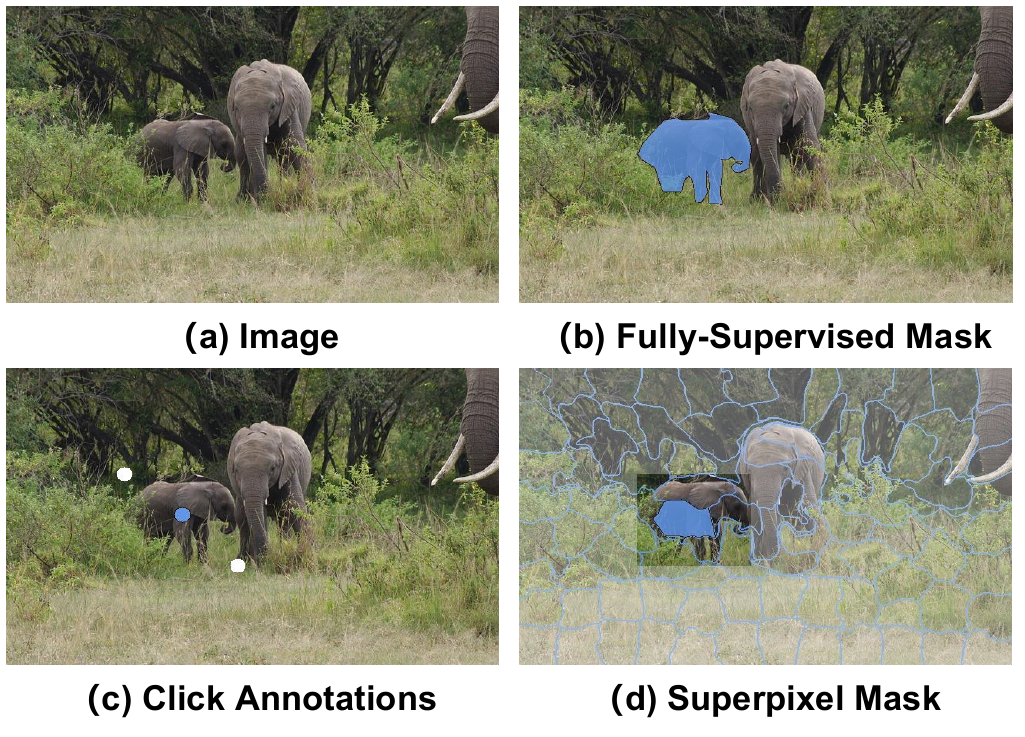}
\end{center}
   \caption{Annotated example for weakly-supervised RES from dataset RefCOCO~\cite{yu2016modeling}. (a) The original image. (b) The fully-supervised mask, where the blue area is the ground-truth mask. (c) Click annotations, where the blue point is the object center point, while the white points are the object corners. (d) Superpixel mask, where the blue area is the superpixel mask, while the areas outside the corner points are the background.
}
\label{fig:data}
\end{figure}

\begin{table}
  \centering
  \caption{Comparison of the cost between fully-supervised annotations and weakly-supervised annotations for one target.The time cost is obtained from~\cite{bearman2016s,everingham2010pascal}.}
  \label{tab:bb}
  \begin{tabular}{l|l|l}
    \hline
    \hline
    Parameters & Fully-Supervised & Weakly-Supervised\\
    \hline
    Request & polygon & points\\
    Number of points & $\approx$ 15 & 3\\
    Time & $\approx$ 79s & $\approx$ 7.2s\\
   \hline
   \hline
\end{tabular}
\end{table}

\begin{figure*}
\centering
\includegraphics[width=0.95\linewidth]{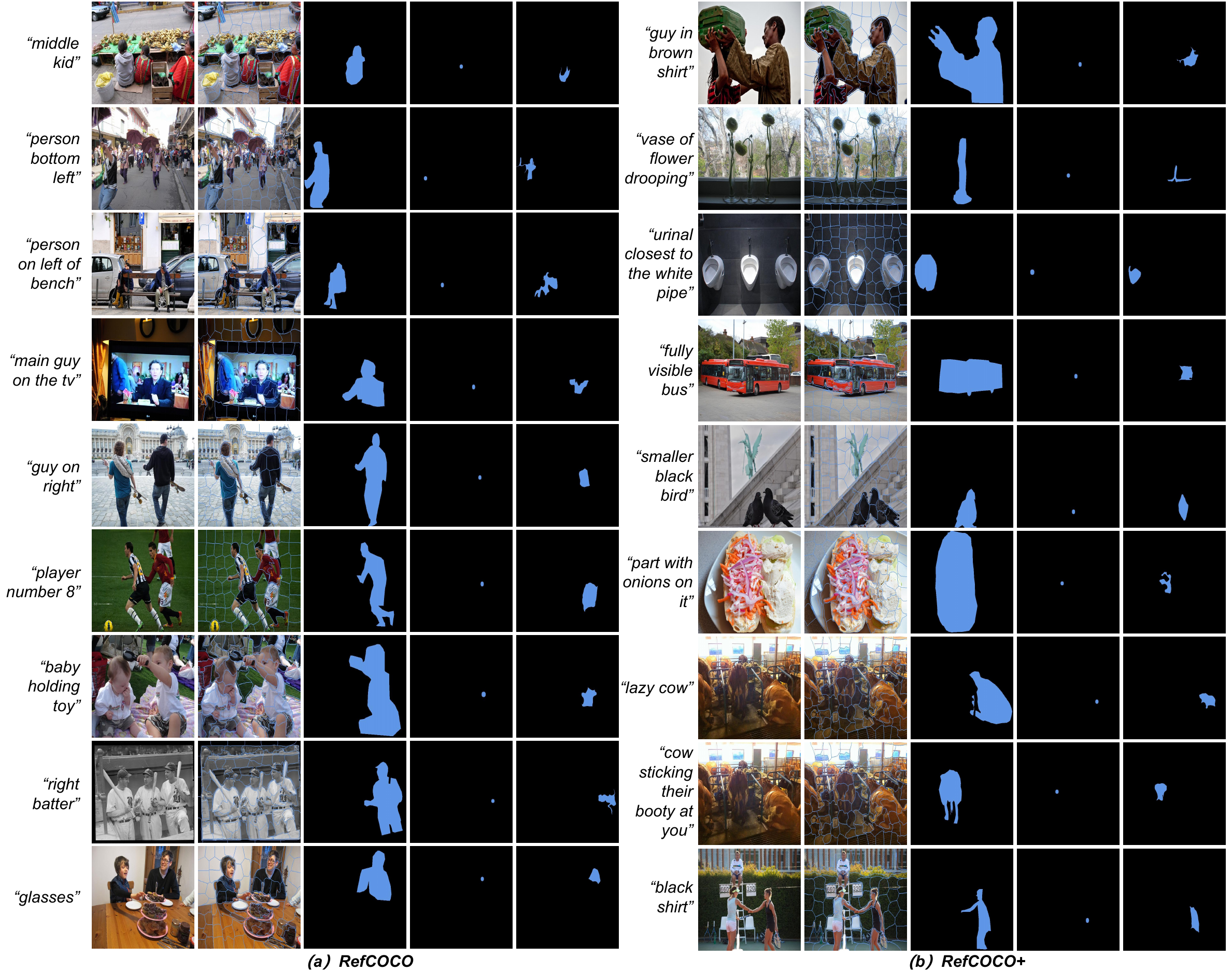}
   \caption{Examples of the fully-supervised and weakly-supervised annotations on (a) RefCOCO~\cite{yu2016modeling} and (b) RefCOCO+~\cite{yu2016modeling}. For each example, we provide: the expression, original image, superpixel pieces, fully-supervised mask, center point annotation, and weakly-supervised superpixel mask.}
\label{fig:weakA}
\end{figure*}

\subsection{Fully-Supervised Losses}
\label{sec:loss}

In the localization branch, the training loss $\mathcal{L}_{pos}$ measures the difference between the predicted position heatmaps $D$ and the ground-truth label:
\begin{equation}
\centering
\label{eq:loss_pos}
\mathcal{L}_{pos} = \frac{1}{L} \sum_{i=1}^{L}\mathit{Focal}(d_i,y^{pos}_i), 
\end{equation}
which uses the focal loss~\cite{lin2017focal} to optimize the predicted $d_i$ on all $L$ levels. 
$y^{pos}_i\in\{0,1\}^{h_i\times w_i}$ is the one-hot ground-truth map for the $i$-th position map, where the object's center $3\times3$ points are set to 1 to represent the position of the target object.


In the segmentation branch, taking the predicted mask $P$ from~Eq.(\ref{eq:mask}), the segmentation loss is:
\begin{equation}
\centering
\label{eq:loss_seg}
\mathcal{L}_{seg} = \mathit{Dice}(P,Y^{seg}), 
\end{equation}
where the Dice loss~\cite{milletari2016v} is adopted to measure the mask difference. $Y^{seg}\in\{0,1\}^{\frac{H}{4}\times \frac{W}{4}}$ is the one-hot ground-truth mask of the target object with an $1/4$ bilinear down-sampling process.

The final training loss is the sum of Eq.(\ref{eq:loss_pos}) and Eq.(\ref{eq:loss_seg}):
\begin{equation}
\centering
\label{eq:loss_all}
\mathcal{L} = \mathcal{L}_{pos} + \mathcal{L}_{seg}. 
\end{equation}

In the inference process, the Matrix NMS~\cite{wang2020solov2} is employed to merge all the predicted masks $P$ from~Eq.(\ref{eq:mask}) into one final segmentation mask.

\begin{table*}[ht]
\centering
 \caption{Comparisons between the proposed method and other SOTA methods according to the mIoU score, on RefCOCO~\cite{yu2016modeling}, RefCOCO+~\cite{yu2016modeling} and RefCOCOg~\cite{mao2016generation,nagaraja2016modeling} datasets. U: UMD split. G: Google split.}
 \label{tab:SOTA}
\begin{tabular}{lc|ccc|ccc|ccc}
\hline
\hline 
\multirow{2}{*}{} & \multirow{2}{*}{Backbone} & \multicolumn{3}{c}{RefCOCO} & \multicolumn{3}{c}{RefCOCO+} & \multicolumn{3}{c}{RefCOCOg} \\
\cline{3-11}
 &  & val & test A & test B & val & test A & test B & val (U) & test (U) & val(G) \\
 \hline
RMI~\cite{liu2017recurrent} \tiny{ICCV'17} & ResNet101 & 45.18 & 45.69 & 45.57 & 29.86 & 30.48 & 29.50 & - & - & 34.52 \\
DMN~\cite{margffoy2018dynamic} \tiny{ECCV'18} & ResNet101 & 49.78 & 54.83 & 45.13 & 38.88 & 44.22 & 32.29 & - & - & 36.76 \\
RRN~\cite{li2018referring}+DCRF \tiny{CVPR'18} & ResNet101 & 55.33 & 57.26 & 53.93 & 39.75 & 42.15 & 36.11 & - & - & 36.45 \\
MAttNet~\cite{yu2018mattnet} \tiny{CVPR'18} & ResNet101 & 56.51 & 62.37 & 51.70 & 46.67 & 52.39 & 40.08 & 47.64 & 48.61 & - \\
CMSA~\cite{ye2019cross}+DCRF \tiny{CVPR'19} & ResNet101 & 58.32 & 60.61 & 55.09 & 43.76 & 47.60 & 37.89 & - & - & 39.98 \\
BRINet~\cite{hu2020bi} \tiny{CVPR'20}& DeepLab-101 & 60.98 & 62.99 & 59.21 & 48.17 & 52.32 & 42.11 & - & - & 48.04 \\
CMPC~\cite{huang2020referring} \tiny{CVPR'20} & DeepLab-101 & 61.36 & 64.53 & 59.64 & 49.56 & 53.44 & 43.23 & - & - & 39.98 \\
LSCM~\cite{hui2020linguistic} \tiny{ECCV'20} & DeepLab-101 & 61.47 & 64.99 & 59.55 & 49.34 & 53.12 & 43.50 & - & - & 48.05 \\
MCN~\cite{luo2020multi} \tiny{CVPR'20} & DarkNet53 & 62.44 & 64.20 & 59.71 & 50.62 & 54.99 & 44.69 & 49.22 & 49.40 & - \\
CGAN~\cite{luo2020cascade} \tiny{ACM MM'20} & DarkNet53 & 64.86 & 68.04 & 62.07 & 51.03 & 55.51 & 44.06 & 51.01 & 51.69 & 46.54 \\
ACM~\cite{feng2021encoder} \tiny{CVPR'21}& ResNet101 & 62.76 & 65.69 & 59.67 & 51.50 & 55.24 & 43.01 & 51.93 & - & - \\
LTS~\cite{jing2021locate} \tiny{CVPR'21} & DarkNet53 & 65.43 & 67.76 & 63.08 & 54.21 & 58.32 & 48.02 & 54.40 & 54.25 & - \\
VLT~\cite{vision-language-transformer} \tiny{ICCV'21} & Transformer & 65.65 & 68.29 & 62.73 & 55.50 & 59.20 & 49.36 & 52.99 & 56.65 & 49.76 \\
ReSTR~\cite{ReSTR} \tiny{CVPR'22} & Transformer & 67.22 & 69.30 & 64.45 & 55.78 & 60.44 & 48.27 & - & - & 54.58 \\

\hline
\rowcolor{lightgray}
 PKS & ResNet101 & \textbf{70.87} & \textbf{74.23} & \textbf{65.07} & \textbf{60.90} & \textbf{66.21} & \textbf{51.35} & \textbf{60.38} & \textbf{60.98} & \textbf{56.97} \\
\hline
\hline
\end{tabular}
\end{table*}

\begin{table}
\centering
\caption{Ablation study of the influence of the kernel head on the RefCOCO~\cite{yu2016modeling} validation set.}
\setlength{\tabcolsep}{1mm}{
\label{tab:kernel}
\begin{tabular}{c|ccccc|c}
\hline
\hline
\multirow{2}{*}{\emph{Modules}} & \multicolumn{5}{c|}{prec@X} & \multirow{2}{*}{mIoU} \\ 
\cline{2-6}
  & 0.5 & 0.6 & 0.7 & 0.8 & 0.9 &\\
 \hline
 w/o kernel & 77.45 &74.98 &70.66 &61.08 &34.01 &68.27 \\
 w/ kernel & \textbf{80.13}& \textbf{77.67}& \textbf{73.15}& \textbf{64.60}& \textbf{35.87}& \textbf{70.87} \\
 \hline
 & \emph{+2.68}& \emph{+2.69}& \emph{+2.49} & \emph{+3.52}& \emph{+1.86} & \emph{+2.60}\\
 \hline
 \hline
\end{tabular}}
\end{table}

\section{Extension for Weakly-Supervised Setting}
\label{sec:weak}
\subsection{Annotation of Weakly-Supervised RES}
\label{sec:weak_annotation}

Fully-supervised RES requests the annotator to draw precious polygon vertices of the target object (Fig.\ref{fig:data}(b)). In contrast, weakly-supervised RES only needs clicks in the labelling process. Fig.\ref{fig:data}(c) shows one example of the weak annotations for the target object ``baby elephant''.

We annotate three datasets including RefCOCO~\cite{yu2016modeling}, RefCOCO+~\cite{yu2016modeling} and RefCOCOg~\cite{mao2016generation,nagaraja2016modeling}. 
For each object, we need three key points for the target object: (1) center of the object; (2) left-top corner of the object, and (3) the right-down corner of the object. 
In our case, we use the object's center-of-mass as the center point, the left-top and right-down bounding box corners as two corner clicks for convenience.

Inspired by the semantic segmentation task, we use superpixel~\cite{7885581} to expand the center point annotation to generate initial labels for the target object. The areas outside the two corners are treated as the background for the target object. The remaining regions are defined as unknown areas (Fig.\ref{fig:data}(d)).

Table~\ref{tab:bb} compares the annotation cost of fully-supervised and weakly-supervised RES. Each target object needs approximately 15 points for a precious boundary, costing 79 s~\cite{bearman2016s}. Our weakly-supervised annotation only needs 3 points for one object, costing 7.2 s, which is approximately ten times faster than the fully-supervised counterpart. A lower annotation cost also means lower financial budget. Some annotation examples are shown in Fig.\ref{fig:weakA}. 

\subsection{Training Losses of Weakly-Supervised RES}
\label{sec:loss_w}

In the localization branch, the training loss $\mathcal{\hat{L}}_{pos}$ measures the difference between the predicted position heatmaps $D$ and the ground-truth label, which involves the same calculation as~Eq.(\ref{eq:loss_pos}).

In the segmentation branch, using the predicted mask $P$ from~Eq.(\ref{eq:mask}), the weakly-supervised segmentation loss is:
\begin{equation}
\centering
\label{eq:loss_seg_w}
\mathcal{\hat{L}}_{seg} = \mathit{pCE}(P,\hat{Y}^{seg}), 
\end{equation}
where partial Cross-Entropy Loss (pCE) is adopted to measure the mask difference. $\hat{Y}^{seg}\in\{0,1,-1\}^{\frac{H}{4}\times \frac{W}{4}}$ is the ground-truth mask of the target object for weakly-supervised RES, which is defined in Sect.\ref{sec:weak_annotation}. We only consider the foreground and background areas of the target object (as 1 and 0) and disregard the unknown areas (as -1) during the training process.

To better consider the color information from the image, we adopt the widely-applied Gated-CRF (GCRF) loss~\cite{obukhov2019gated} in the training process: 
\begin{equation}
\centering
\label{eq:loss_CRF_w}
\mathcal{\hat{L}}_{w} = \mathit{GCRF}(P,\hat{I}), 
\end{equation}
where $\hat{I}$ is the resized raw RGB image and $P$ is the predicted mask. 

The final training loss for weakly-supervised RES is:
\begin{equation}
\centering
\label{eq:loss_all_w}
\mathcal{\hat{L}} = \mathcal{\hat{L}}_{pos} + \mathcal{\hat{L}}_{seg} + \lambda_{w} \mathcal{\hat{L}}_{w},
\end{equation}
where $\lambda_{w}$ is the hyperparameters to balance the influence of the GCRF loss.

\section{Experiments}

\subsection{Experimental Settings}
\textbf{Metrics:}
The RES has two evaluation metrics: the mean Intersection over Union (mIoU) and the precision score over a threshold X (prec@X). mIoU calculates intersection regions over union regions of the predicted segmentation mask and the ground-truth. prec@X measures the percentage of test images with an IoU score higher than the threshold X , where X$\in \{0.5,0.6,0.7,0.8,0.9\}$.

\begin{table*}
\centering
\caption{Ablation study of different fusion strategies, conducted on the validation set of RefCOCO.}
\label{tab:AB_com}
\begin{tabular}{lcc|ccccc|c}
\hline
\hline
\multicolumn{3}{c|}{\emph{modules}} & \multicolumn{5}{c|}{prec@X} & \multirow{2}{*}{mIoU} \\
\cline{1-8}
 \emph{strategy} & \emph{multi-level} &\emph{multi-step} & prec@0.5 & prec@0.6 & prec@0.7 & prec@0.8 & prec@0.9 &  \\
 \hline
   concat & \ding{51} & \ding{51} & 33.29  &32.63  &31.41  & 28.50 &17.10  &29.97  \\
  multipy   &\ding{51}&  \ding{51}   & 64.91 & 60.78 & 55.73 & 46.17 & 22.10 & 56.76\\
  mean   &\ding{51} &  \ding{51}   & 72.67& 69.58& 64.58& 55.21& 28.61& 63.89   \\
  \hline
 MSF & \ding{55} & \ding{55} & 69.37  &67.16  &63.18  & 55.99 &31.23  &61.61  \\
  MSF   &\ding{51}&  \ding{55}   & 76.28 & 73.83 & 69.89 & 61.75 & 34.88 & 67.52\\
  MSF   &\ding{55} &  \ding{51}   & 76.33& 73.36& 68.82& 60.50& 33.97& 67.43   \\
  \hline
  MSF   &\ding{51}  & \ding{51}    & \textbf{80.13}& \textbf{77.67}& \textbf{73.15}& \textbf{64.60}& \textbf{35.87}& \textbf{70.87}  \\
 \hline
 \hline
\end{tabular}
\end{table*}

\begin{table}
\centering
\caption{Analysis on how the step number $J$ affects the performance, conducted on the validation set of RefCOCO~\cite{yu2016modeling}. The results are evaluated by prec@X and mIoU, on different settings of $J$ with/without Multi-Level (ML) features.}
\setlength{\tabcolsep}{1.2mm}{
\label{tab:AB_ms}
\begin{tabular}{cc|ccccc|c}
\hline
\hline
\multirow{2}{*}{\emph{ML}} & \multirow{2}{*}{\emph{$J$}} & \multicolumn{5}{c|}{prec@X} & \multirow{2}{*}{mIoU} \\
\cline{3-7}
  && 0.5 & 0.6 & 0.7 & 0.8 & 0.9 &\\
 \hline
 \multirow{5}{*}{\ding{55}}
 &1 &69.37  &67.16  &63.18  & 55.99 &31.23  &61.61  \\
 &2& 73.76& 71.05& 67.20 & 59.35 & 32.19 & 65.47\\
 &3& 76.14 & 73.07 & \textbf{69.01} & \textbf{60.74} & \textbf{34.01} & 67.39  \\
 &4& \textbf{76.33}& \textbf{73.36}& 68.82& 60.50& 33.97& \textbf{67.43} \\
 \hline
 \multirow{5}{*}{\ding{51}}
 &1 &76.28 & 73.83 & 69.89 & 61.75 & 34.88 & 67.52  \\
 &2& 79.21 & 75.43 & 72.10 & 63.09 & 35.58 & 69.72 \\
 &3& 79.87 & 77.19 & 72.88 & 63.93 & 35.85 & 70.58 \\
 &4& \textbf{80.13}& \textbf{77.67}& \textbf{73.15}& \textbf{64.60}& \textbf{35.87}& \textbf{70.87} \\
 \hline
 \hline
\end{tabular}}
\end{table}

\textbf{Dataset:}
The proposed method is evaluated on three main RES datasets, including RefCOCO~\cite{yu2016modeling}, RefCOCO+~\cite{yu2016modeling} and RefCOCOg~\cite{mao2016generation,nagaraja2016modeling}.

RefCOCO is a standard RES dataset, where the numbers of images, objects, and expressions are 19,994, 50,000, and 142,210, respectively. Expressions are split into 120,624, 10,834, 5,657, and 5,095, as the train, val, testA, and testB sets, respectively. A single image contains multiple targets, each of which is annotated by several sentences.  

RefCOCO+ is another widely-employed RES dataset, where fewer spatial descriptions are applied in referring expressions. The numbers of images, objects, and expressions are 19,992, 49,856, and 141,564, respectively. Expressions are split into 120,191, 10,758, 5,726, and 4,889,  as the train, val, testA, and testB sets, respectively.

RefCOCOg has two different branches: Google~\cite{mao2016generation} and UMD~\cite{nagaraja2016modeling}. In terms of Google, the numbers of images, objects, and expressions are 26,711, 54,822, and 85,474, respectively. Objects are split into 44,822, 5,000, and 5,000 as train, val, and test sets, respectively, where the test set is not released. For UMD, the numbers of images, objects, and expressions are 25,799, 49,822, and 95,010, respectively. Objects are split into 42,226, 2,573, and 5,023 as the train, val, test and sets, respectively.

\subsection{Implementation Details}
The adopted visual feature extractor is ResNet-101 \cite{he2016deep}, where the feature maps of all levels in FPN~\cite{lin2017feature} are used as the multiscale inputs of the later modules. The linguistic feature extractor is Bi-GRU~\cite{cho2014properties}, where a sequential bidirectional module is utilized to embed the sentence and each word is embedded into a 1024-D vector. In terms of the MSF module described in~Sect.\ref{sec:position}, we set the step number $J$=$4$ in Eq.(\ref{eq:MS1}). The hyperparameter $\lambda_f$ for kernel selection in Eq.(\ref{eq:filter}) is set to 0.1, as in SOLOv2~\cite{wang2020solov2}.

The training process is conducted on four 32G TESLA V100 GPUs, with a batch size of 8 for each GPU. The process takes approximately 50 hours for 30 epochs. The input image sizes $H$ and $W$ for all images are rescaled to 1,333 and 800, respectively. SGD is adopted as the optimizer, with the learning rate set as 0.01, the momentum set as 0.9, and the weight decay is set as $1\times 10^{-4}$. All the obtained results have some fluctuations, from -0.2 to 0.2, when different seeds are employed.

For weakly-supervised RES in Sect.\ref{sec:weak}, SLIC~\cite{7885581} is adopted as the superpixel generation method for weakly-supervised annotation, where the number of segmentation areas for each image is set to 100, and the other parameters follow the same settings as in the original paper.
The hyperparameter $\lambda_w$ in Eq.(\ref{eq:loss_all_w}) is set to 0.01. 
The parameters in GCRF loss are the same as in~\cite{obukhov2019gated}. Other settings remain the same as those in the fully-supervised case.

\begin{table}
\centering
\caption{Analysis on position predictions on different range of the feature levels conducted on the validation set of RefCOCO~\cite{yu2016modeling}. The results are evaluated according to prec@X and mIoU. A larger level number represents a deeper feature, with a smaller resolution.}
\setlength{\tabcolsep}{1.2mm}{
\label{tab:AB_ml}
\begin{tabular}{l|ccccc|c}
\hline
\hline
\multirow{2}{*}{\emph{level range}} & \multicolumn{5}{c|}{prec@X} & \multirow{2}{*}{mIoU} \\ 
\cline{2-6}
  & 0.5 & 0.6 & 0.7 & 0.8 & 0.9 &\\
 \hline
 \emph{level 5} & 76.33& 73.36& 68.82& 60.50& 33.97& 67.43 \\
 \emph{level 4-5}& 78.39& 75.89& 71.61 & 62.68 & 34.77 & 69.46\\
 \emph{level 3-5}& 78.98& 76.09& 71.61 & 62.95 & 35.30 & 69.87\\
 \emph{level 2-5}& 79.31& 77.05& 72.22 & 63.85 & 35.34 & 70.18\\
 \emph{level 1-5}& \textbf{80.13}& \textbf{77.67}& \textbf{73.15} & \textbf{64.60}& \textbf{35.87} & \textbf{70.87}\\
 \hline
 \hline
\end{tabular}}
\end{table}

\subsection{Experimental Results for Fully-Supervised RES}

\subsubsection{Comparison with Fully-Supervised SOTA Methods}
Table \ref{tab:SOTA} reports the comparisons between the proposed method and other SOTA methods according to the mIoU score. The proposed method achieves a new SOTA accuracy with a significant gain on three RES datasets, including RefCOCO, RefCOCO+ and RefCOCOg. Compared with the previous SOTA method ReSTR~\cite{ReSTR}, the proposed method improves the mIoU scores (averaged over val, test A and test B) by 3.07\%, 4.66\% and 2.39\% on RefCOCO, RefCOCO+ and RefCOCOg, respectively. This result demonstrates the effectiveness of the proposed parallel pipeline in the RES task. Note that the proposed model is a one-stage framework trained end-to-end without generating proposals or masks in advance. In our method, we use the original ResNet101 as the feature extractor which is the same as~\cite{liu2017recurrent,margffoy2018dynamic,li2018referring,yu2018mattnet,ye2019cross,feng2021encoder}. 
Some other methods use more powerful backbones (DarkNet53, DeepLab-101 and Transformer). Nevertheless, our method still achieves better performances than them, proving the effectiveness of our method.

\begin{table}
\centering
\caption{Upper-bound analysis on the RefCOCO~\cite{yu2016modeling} validation set. \emph{gt pos.} indicates that the ground-truth position of the target object is utilized on the kernel selecting process in~Sect.\ref{sec:kernel}}
\setlength{\tabcolsep}{1mm}{
\label{tab:AB_upp}
\begin{tabular}{c|ccccc|c}
\hline
\hline
\multirow{2}{*}{\emph{gt pos.}} & \multicolumn{5}{c|}{prec@X} & \multirow{2}{*}{mIoU} \\ 
\cline{2-6}
  & 0.5 & 0.6 & 0.7 & 0.8 & 0.9 &\\
 \hline
 \ding{55} &80.13 &77.67 &73.15 &64.60 &35.87 &70.87 \\
 \ding{51} & \textbf{93.28}& \textbf{90.28}& \textbf{84.65} & \textbf{74.50}& \textbf{41.56} & \textbf{81.91} \\
 \hline
 & \emph{+13.15}& \emph{+12.61}& \emph{+11.50} & \emph{+9.90}& \emph{+5.69} & \emph{+11.04}\\
 \hline
 \hline
\end{tabular}}
\end{table}

\begin{figure*}
\centering
\includegraphics[width=0.95\linewidth]{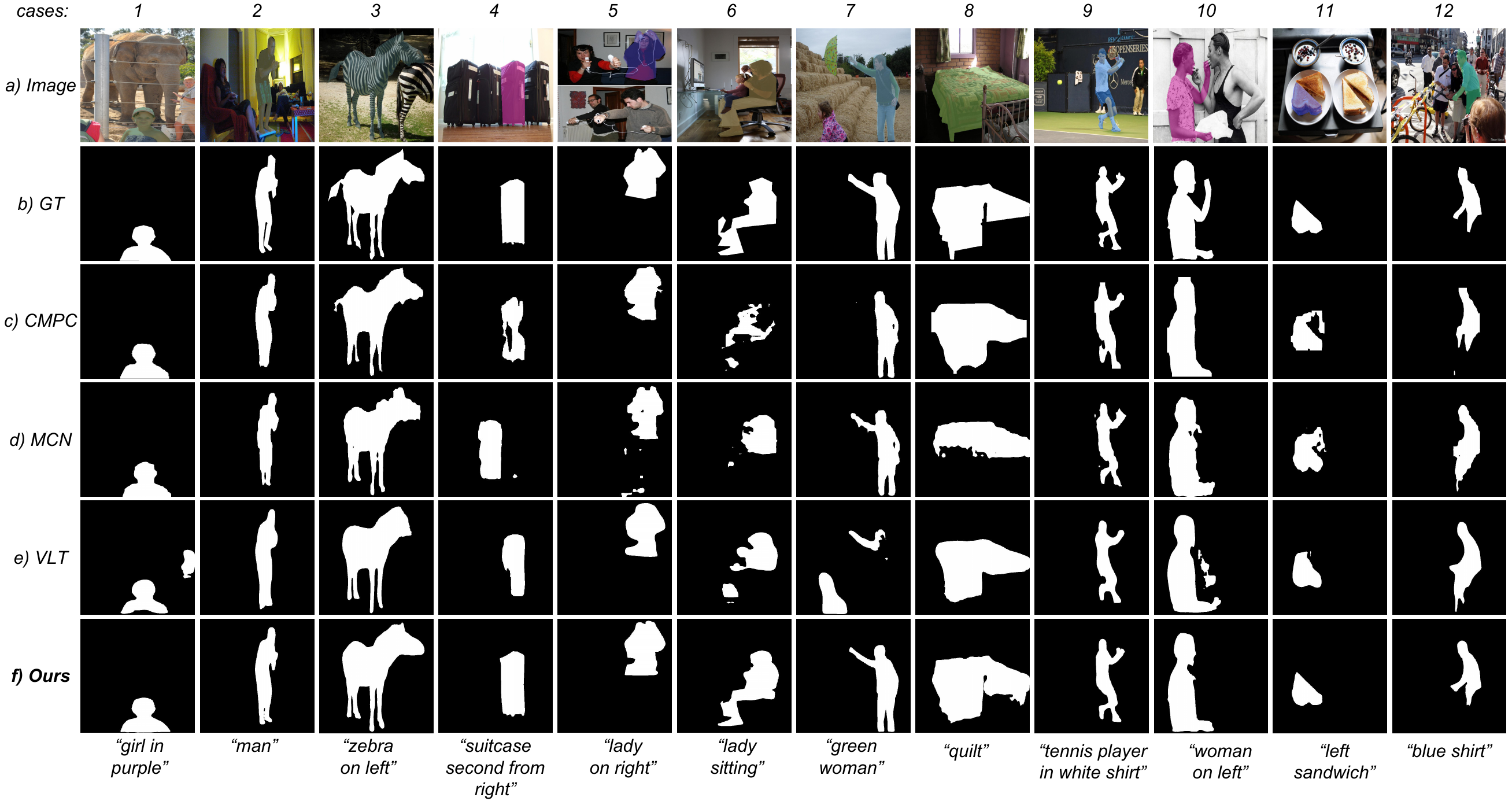}
   \caption{Qualitative comparisons between (f) our method and other SOTA methods: (c) CMPC~\cite{huang2020referring}, (d) MCN~\cite{luo2020multi} and (e) VLT~\cite{vision-language-transformer}, on the validation set of RefCOCO~\cite{yu2016modeling}. The segmentation results are obtained based on the expression at the bottom of each column. Note that (a) Image indicates the original image and (b) GT indicates the ground-truth segmentation mask.}
\label{fig:results}
\end{figure*}

\subsubsection{Ablation Studies on Fully-Supervised Components}
The first ablation study is to show the effectiveness of isolating localization with segmentation, as reported in Table ~\ref{tab:kernel}. Item ``w/o kernel'' means that the attention-fused language-visual features are directly provided to the segmentation predictions (similar to~\cite{jing2021locate,LI202299}). Item ``w/ kernel'' means that the kernel head is introduced as the bridge to isolate the position head and segmentation head, which is our method. The final mIoU score on the validation set of RefCOCO increases from 68.27\% to 70.87\%, with a gain of 2.60\%, proving the effectiveness of isolating the two modules via the kernel head.

The second ablation study is on MSF. This study is conducted on the validation set of RefCOCO, and the results are reported in Table \ref{tab:AB_com}. We implement three different language fusion strategies in addition to our MSF module. ``concat'' means that the language and visual features are simultaneously concatenated on the channel dimension, followed by one FCL to keep the same dimension size in each layer. ``multipy'' means that the language and visual features are element-wise and simultaneously multiplied on the channel dimension, followed by one FCL. ``mean'' represents that the mean feature value of all words is used to present the entire sentence in our MSF. The results prove that our MSF strategy is the most effective feature fusion solution for our method.

We also explore the influence of each component in the proposed method (\emph{i.e.,} multi-level and multi-step). 
When the ``multi-step'' part is disabled, the mIoU score descends from 70.87\% to 67.52\%. The result demonstrates that ``multi-step'', which allocates different phrases with different attention values in different steps, renders language-vision fusion more effective than the traditional one-step fusion process. 
On the other hand, when the ``multi-level'' part is forbidden, the mIoU score declines by 3.44\%, indicating that comprehensively using the visual features extracted under different resolutions can better localize the target objects, especially for the minute features. 

The third ablation study is examines how the number of steps $J$ in Eq.(\ref{eq:MS1}) affects the segmentation performance of the proposed method. As shown in Table \ref{tab:AB_ms}, the proposed method achieves the mIoU score of 61.61\% when $J$=$1$ without multi-level features. The performance is improved when the number of steps increases, demonstrating the necessity of the MSF mechanism. When $J$ increases from 3 to 4, the improvement is limited, thus $J$=$4$ is selected to balance performance and complexity. A similar accuracy change is observed when multi-level features are adopted, indicating the robustness of the proposed method.

In the last ablation study, we evaluate the performance when visual features with a certain range of levels are adopted. During the training stage, only the features of the selected levels are used, while in the inference period, Matrix NMS~\cite{wang2020solov2} is conducted on the predictions of the selected levels to localize the target object. As shown in~Table \ref{tab:AB_ml}, the accuracy improves when more levels of the features are adopted in the position head. The highest mIoU score of 70.87\% is achieved when all levels (``level 1-5'') of features are adopted, showing that visual features of all levels jointly contribute to localizing the target objects.

\begin{table*}
\centering
 \caption{Comparisons between the proposed method and other methods according to the mIoU score, on RefCOCO~\cite{yu2016modeling}, RefCOCO+~\cite{yu2016modeling} and RefCOCOg~\cite{mao2016generation,nagaraja2016modeling} datasets. \textbf{F} represents fully-supervised results, and \textbf{W} represents weakly-supervised results.}
 \label{tab:SOTA_W}
\begin{tabular}{lcccc|ccc|ccc}
\hline
\hline
\multirow{2}{*}{} & \multirow{2}{*}{} & \multicolumn{3}{c}{RefCOCO} & \multicolumn{3}{c}{RefCOCO+} & \multicolumn{3}{c}{RefCOCOg} \\
\cline{3-11}
 & & val & test A & test B & val & test A & test B & val (U) & test (U) & val(G) \\
 \hline
VLT~\cite{vision-language-transformer} & F & 65.65 & 68.29 & 62.73 & 55.50 & 59.20 & 49.36 & 52.99 & 56.65 & 49.76 \\
PKS & F & \textbf{70.87} & \textbf{74.23} & \textbf{65.07} & \textbf{60.90} & \textbf{66.21} & \textbf{51.35} & \textbf{60.38} & \textbf{60.98} & \textbf{56.97} \\
\hline
VLT~\cite{vision-language-transformer} & W & 45.23 & 46.32 & 44.57 & 31.74 & 34.83 & 27.43 & 33.20 & 32.63 & 30.67 \\
\rowcolor{lightgray}
PKS & W & \textbf{49.27} & \textbf{52.23} & \textbf{45.64} & \textbf{37.79} & \textbf{42.09} & \textbf{32.87} & \textbf{40.98} & \textbf{40.80} & \textbf{36.43} \\

\hline
\hline
\end{tabular}
\end{table*}

\begin{table*}
\centering
\caption{Ablation studies on the influence of labels and GCRF loss, for the weakly-supervised setting, conducted on the validation set of RefCOCO.}
\label{tab:AB_weak}
\begin{tabular}{l|ccccccccc|c}
\hline
\hline
\multirow{2}{*}{\emph{modules}} & \multicolumn{9}{c|}{prec@X} & \multirow{2}{*}{mIoU} \\
\cline{2-10}
  & p@0.1 & p@0.2 & p@0.3 & p@0.4 & p@0.5 & p@0.6 & p@0.7 & p@0.8 & p@0.9 &  \\
  \hline
 Fully-Supervised & 85.40  &84.54  &83.57 & 82.12 & 80.13  &77.67  &73.15 & 64.60 & 35.87 & 70.87  \\
 \hline
 baseline + point FG & 28.24  &14.01  &5.96 & 2.04 & 0.69  &0.16  &0.05 & 0.00 & 0.00  & 7.30  \\
 baseline + point FG + BG & 73.35  &57.18  &29.40 & 9.42 & 2.02  &0.35  &0.03 & 0.00 & 0.00  & 21.11  \\
  baseline + superpixel FG + BG & 76.98  &74.07  &70.50 & 64.80& 56.21 & 42.47 & 25.61 & 8.60 & 0.75 & 46.25\\
   \hline
  baseline + superpixel FG + BG + GCRF loss  & \textbf{80.18}  &\textbf{77.90}  &\textbf{74.77} & \textbf{69.69}  & \textbf{61.53}& \textbf{48.35}& \textbf{28.56}& \textbf{9.56}& \textbf{0.78}& \textbf{49.27}  \\
 \hline
 \hline
\end{tabular}
\end{table*}

\begin{figure}
\centering
\includegraphics[width=0.95\linewidth]{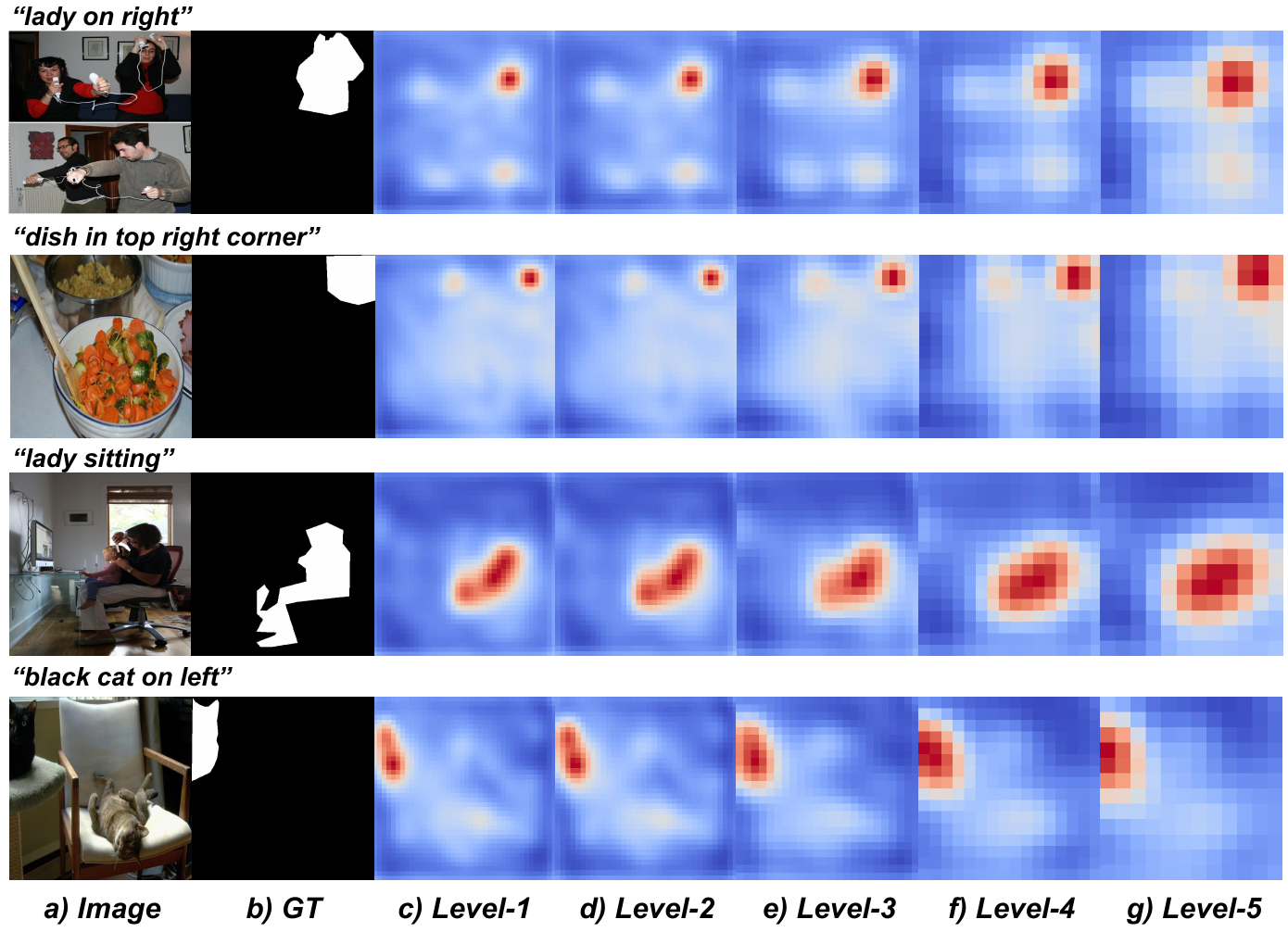}
   \caption{Visualization of the position heatmaps on different feature levels, where red areas represent the target regions while blue areas indicate the non-target regions. A larger level number represents a deeper feature, with a smaller resolution.}
\label{fig:ML}
\end{figure}

\subsubsection{Analysis on Upper-bound}

We also analyse the upper-bound of our method conducted on the validation set of RefCOCO in~Table \ref{tab:AB_upp}. The upper-bound denotes that the ground-truth position of the target object is utilized in the kernel selection process in~Sect.\ref{sec:kernel}. The performance difference between the upper-bound and the standard setting in our method on position is 13.15\% at prec@0.5 (prec@0.5 is the normal overlap threshold of correct findings). Even with the ground-truth position, the proposed method achieves an mIoU of 81.91\%, where the predicted masks still do not fully overlap the ground-truth masks. 

\subsubsection{Qualitative Analysis for Fully-Supervised RES}
The qualitative comparison between our method and other SOTA methods, including CMPC~\cite{huang2020referring}, VLT~\cite{vision-language-transformer}, and MCN~\cite{luo2020multi}, is shown in~Fig.\ref{fig:results}. Different from the broken masks predicted by CMPC~\cite{huang2020referring}, the proposed method generates more intact results, such as ``suitcase'' in Case (4) and ``green woman'' in Case (7). Compared with MCN~\cite{luo2020multi}, 
our proposed method predicts better contours, such as ``lady sitting'' in Case (6)  and ``quilt'' in Case (8). Compared with VLT~\cite{vision-language-transformer}, 
whose prediction covers other distracting objects similar to the target, the proposed method correctly distinguishes the target from other distracting objects, such as ``girl in purple'' in Case (1)  and ``green woman'' in Case (7).

The visualization of the position heatmaps corresponding to different levels is provided in~Fig.\ref{fig:ML}. As observed, the target center is adequately localized in all heatmaps. The low-resolution deep-level features (such as level-5), which contains more semantic information, take more responsibility for identifying the main region of the object. However, the low-level feature, with a larger resolution, can predict a more precise position heatmap, especially for irregular, long-shaped objects. In this way, all these features are complementary, which jointly contributes to the localization of the target.

\begin{figure}
\centering
\includegraphics[width=0.8\linewidth]{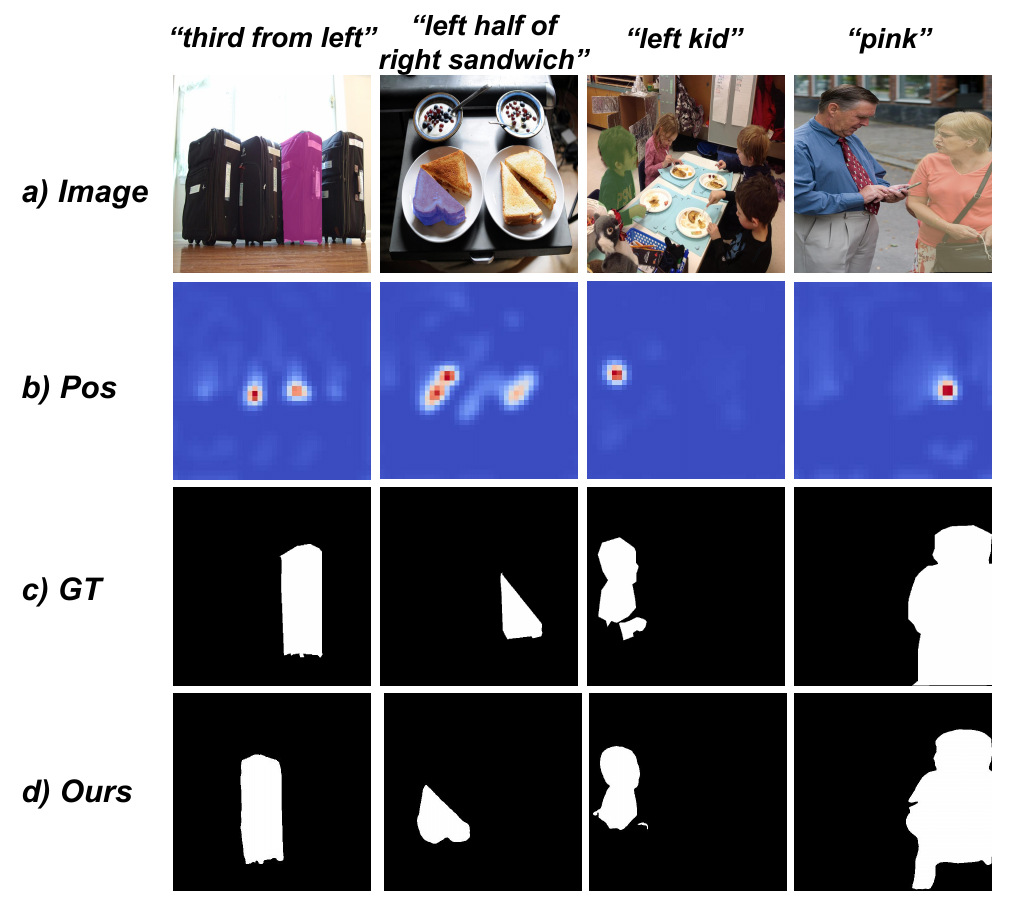}
   \caption{Visualization of failed cases on the validation set of RefCOCO~\cite{yu2016modeling}. Note that GT indicates the ground truth segmentation mask and Pos indicates the position prediction. Best viewed in colors. }
\label{fig:fail}
\end{figure}

In addition, four failed cases are illustrated in~Fig.\ref{fig:fail}. In the left two cases, the target object is not properly localized, and the distracting object is predicted. Specifically, the first case shows that it is arduous for the proposed method to handle ordinal numbers (e.g., ``third''), as these words are not common in the training set. In the second case, the proposed method fails as too many spatial descriptions occur in a single referring sentence (e.g., ``left half'' and ``right''). The right two cases show that the segmentation quality needs to be further improved, especially when the target object contour is hard to identify, or the target object consists of several separate parts.

\begin{figure*}
\centering
\includegraphics[width=0.95\linewidth]{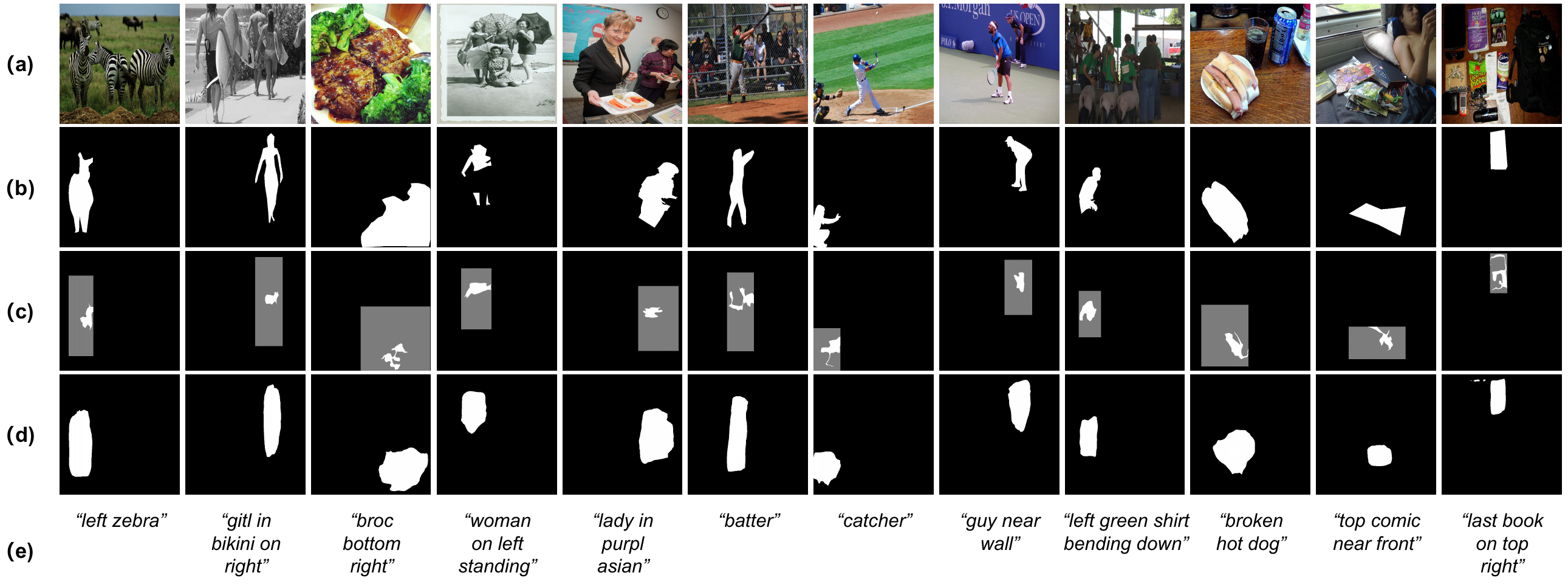}
   \caption{Qualitative results of weakly-supervised RES on the validation set of RefCOCO. (a) Original image. (b) Fully-supervised masks, white regions are the foreground and black regions are the background. (c) Weakly-supervised masks, the grey regions have unknown labels for the training process. (d) Weakly-supervised predictions.}
\label{fig:weakB}
\end{figure*}

\subsection{Experimental Results for Weakly-Supervised Setting}
\subsubsection{Comparison with Other Methods}

Table \ref{tab:SOTA_W} compares the mIoU scores between the proposed method and VLT~\cite{vision-language-transformer}, on both fully-supervised (F) and weakly-supervised (W) RES. The fully-supervised VLT is trained with the originally published code, and the weakly-supervised VLT is trained with superpixel masks by replacing CE loss with pCE loss. 
As observed, our method achieves clear gains for both fully-supervised and weakly-supervised settings. 

\subsubsection{Ablation Studies on a Weakly-Supervised Setting}

Table \ref{tab:AB_weak} reports the performances when some components are disabled, to show their effectiveness.  ``Point FG'' represents the setting that the ground-truth labels only contain the center point as the foreground. ``Point FG + BG'' means that the ground-truth labels contain both the foreground points and background points. ``superpixel FG + BG'' means that the superpixel method is used to expand the foreground labels to replace the center point foreground. ``superpixel FG + BG + GCRF loss'' represents that the training optimization adopts the GCRF loss. As observed, with only the center point as the foreground, a mIoU score of 7.30\% is achieved. After adding the background points, the mIoU score increases to 21.11\%, showing that the foreground labels do not contain enough information and that background annotation is necessary for weakly-supervised RES training process. When replacing the center point with the superpixel foreground, a major improvement from 21.11\% to 46.25\% is obtained. After adopting the GCRF loss, the mIoU score increases to 49.27\%. We observe that the gap between the fully-supervised and weakly-supervised results is 21.60\%, indicating that there is still room for further improvement for weakly-supervised RES.

\subsubsection{Qualitative Analysis for Weakly-Supervised RES}

Fig.\ref{fig:weakB} provides examples of the weakly-supervised RES results. The target objects are sufficiently localized based on the expression, which shows that our position head is performed well with the click annotation as supervision. However, the predicted masks are not precise, and the details of the objects are missing, due to a lack of detailed ground-truth masks. We suggest that the segmentation head could be further improved, especially when the target object shape is complex. For instance, object saliency information could be used to improve the object boundary, which will be addressed in future research. 


\section{Conclusion}

In this paper, we have individually analysed the two main components of referring expression segmentation: localization and segmentation. We propose a position-kernel-segmentation framework to train the localization and segmentation process in parallel and then interact the localization results with the segmentation process via a visual kernel. Our parallel framework could generate complete, smooth and precise masks, achieving state-of-the-art performances with more than 4.5\% mIoU gain over previous SOTA methods on three main RES datasets. Our framework also enables training RES in a weakly-supervised way, where the position head is fully-supervised and trained with the click annotations as supervision, and the segmentation head is trained with weak segmentation losses. We annotate three benchmark datasets for weakly-supervised training, and achieve satisfactory performances.




\ifCLASSOPTIONcaptionsoff
  \newpage
\fi



\bibliographystyle{IEEEtran}
%
\bibliography{ref}

\end{document}